\documentclass{article}
\usepackage{arxiv}
\usepackage{appendix}
\usepackage{cite}
\usepackage{amsmath,amssymb,amsfonts}
\usepackage{algorithmic}
\usepackage{graphicx}
\usepackage{textcomp}
\usepackage{xcolor}
\usepackage{pgfplots}
\pgfplotsset{compat=1.18} 
\usepackage{xcolor}
\usepackage{listings}
\usepackage[frozencache,cachedir=./]{minted}
\usepackage{tcolorbox}
\usepackage{pdfpages}

\usepackage{xspace}
\usepackage{moresize}

\tcbuselibrary{minted}

\usepackage[utf8]{inputenc} %
\usepackage[T1]{fontenc}    %
\usepackage{hyperref}       %
\usepackage{url}            %
\usepackage{booktabs}       %
\usepackage{amsfonts}       %
\usepackage{nicefrac}       %
\usepackage{microtype}      %
\usepackage{cleveref}       %
\usepackage{lipsum}         %
\usepackage{graphicx}
\usepackage{doi}

\title{Neuro Symbolic Reasoning for Planning: Counterexample Guided Inductive Synthesis using Large Language Models and Satisfiability Solving}

\date{}

\newif\ifuniqueAffiliation
\uniqueAffiliationtrue

\ifuniqueAffiliation %
\author{ 
    {\href{https://orcid.org/0000-0003-0354-2940}{\includegraphics[scale=0.06]{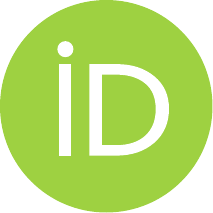}}\hspace{1mm}Sumit Kumar Jha}\\
    Computer Science Department\\
    Florida International University\\
    \texttt{sumit.jha@fiu.edu} \\
    \And
    {\href{https://orcid.org/0000-0001-5983-9095}{\includegraphics[scale=0.06]{orcid.pdf}}\hspace{1mm}Susmit Jha}\\
    Computer Science Laboratory\\
    SRI International\\
    \texttt{susmit.jha@sri.com} \\
    \And
    {Patrick Lincoln} \\
    Computer Science Laboratory\\
    SRI International\\
\texttt{patrick.lincoln@sri.com} \\
    \And
    {\href{https://orcid.org/0000-0001-9957-2778}{\includegraphics[scale=0.06]{orcid.pdf}}\hspace{1mm}Nathaniel D. Bastian} \\
    Army Cyber Institute\\
    United States Military Academy\\
    \texttt{nathaniel.bastian@westpoint.edu} \\
    \And
    {\href{https://orcid.org/0000-0001-6757-105X}{\includegraphics[scale=0.06]{orcid.pdf}}\hspace{1mm}Alvaro Velasquez} \\
    Department of Computer Science\\
    University of Colorado Boulder\\
    \texttt{alvaro.velasquez@colorado.edu} \\
    \And
    {Rickard Ewetz} \\
    Electrical and Computer Engineering \\
    University of Central Florida\\
    \texttt{rickard.ewetz@ucf.edu} \\
    \And
    {Sandeep Neema} \\
    Electrical and Computer Engineering \\
    Vanderbilt University\\
    \texttt{sandeep.neema@vanderbilt.edu} \\
}
\else
\usepackage{authblk}

\newbox{\orcid}\sbox{\orcid}{\includegraphics[scale=0.06]{orcid.pdf}} 
\author[1]{%
	\href{https://orcid.org/0000-0000-0000-0000}{\usebox{\orcid}\hspace{1mm}David S.~Hippocampus\thanks{\texttt{hippo@cs.cranberry-lemon.edu}}}%
}
\author[1,2]{%
	\href{https://orcid.org/0000-0000-0000-0000}{\usebox{\orcid}\hspace{1mm}Elias D.~Striatum\thanks{\texttt{stariate@ee.mount-sheikh.edu}}}%
}
\affil[1]{Department of Computer Science, Cranberry-Lemon University, Pittsburgh, PA 15213}
\affil[2]{Department of Electrical Engineering, Mount-Sheikh University, Santa Narimana, Levand}
\fi

\renewcommand{\shorttitle}{Planning using Neuro Symbolic Reasoning}

\hypersetup{
pdftitle={Planning using Neuro Symbolic Reasoning},
pdfsubject={cs.LG, cs.AI},
pdfauthor={Sumit Kumar Jha},
pdfkeywords={Robust Learning, Planning, Program Synthesis, Large Language Model, Decision Procedures, Satisfiability Solving},
}

\newcommand{\class}{\ensuremath{\mathcal{C}}\xspace}
\newcommand{\concept}{\ensuremath{\mathit{c}}\xspace}

\newcommand{\Spec}{\ensuremath{\Phi}\xspace}
\newcommand{\target}{\ensuremath{\mathit{c}}\xspace}
\newcommand{\domain}{\ensuremath{\mathbf{E}}\xspace}
\newcommand{\ex}{\ensuremath{\mathit{x}}\xspace}

\newcommand{\orint}{\ensuremath{\mathcal{O}}}

\newcommand{\defquery}{\ensuremath{\mathcal{Q}}}
\newcommand{\defresponse}{\ensuremath{\mathcal{R}}}

\newcommand{\powset}[1]{\ensuremath{2^{#1}}}

\newtcbox{\codebox}[1][green!60!black]{on line, boxrule=0.5pt, colback=white, colframe=#1, fontupper=\small\ttfamily}

\begin{document}

\maketitle

\begin{abstract}
Generative large language models (LLMs) with instruct training such as GPT-4 can follow human-provided instruction prompts and generate human-like responses to these prompts. Apart from natural language responses, they have also been found to be effective at generating formal artifacts such as code, plans, and logical specifications from natural language prompts. Despite their remarkably improved accuracy, these models are still known to produce factually incorrect or contextually inappropriate results despite their syntactic coherence -- a phenomenon often referred to as {\em hallucinations}. This limitation makes it difficult to use these models to synthesize formal artifacts that are used in safety-critical applications. Unlike tasks such as text summarization and question-answering, bugs in code, plan, and other formal artifacts produced by LLMs can be catastrophic. We posit that we can use the satisfiability modulo theory (SMT) solvers as deductive reasoning engines to analyze the generated solutions from the LLMs, produce counterexamples when the solutions are incorrect, and provide that feedback to the LLMs exploiting the dialog capability of instruct-trained LLMs. This interaction between inductive LLMs and deductive SMT solvers can iteratively steer the LLM to generate the correct response. 
In our experiments, we use planning over the domain of blocks as our synthesis task for evaluating our approach. We use GPT-4, GPT3.5 Turbo, Davinci, Curie, Babbage, and Ada as the LLMs and Z3 as the SMT solver. Our method allows the user to communicate the planning problem in natural language; even the formulation of queries to SMT solvers is automatically generated from natural language. Thus, the proposed technique can enable non-expert users to describe their problems in natural language, and the combination of LLMs and SMT solvers can produce provably correct solutions. 

\end{abstract}


\section{Introduction}
Many safety-critical applications such as communication protocols, adaptive software-defined radio, autonomous vehicles, and assistive robots need to operate in the presence of uncertainty and have the capability to adapt to an evolving non-stationary environment. This need for generalization and adaptation has fueled the adoption of artificial intelligence (AI) for such applications. We can categorize these AI components into two categories - inductive learning-based methods that generalize well but do not provide any guarantees on their output, and deductive automated reasoning methods that require significant user expertise but provide guarantees on their outputs. Examples of inductive learning methods include decision trees~\cite{kotsiantis2013decision}, support vector machines~\cite{hearst1998support}, deep learning models~\cite{goodfellow2016deep}, the more recent class of large language models (LLMs), and, more generally, foundation models~\cite{bommasani2021opportunities} that generalize across tasks and domains. Examples of deductive symbolic reasoning methods include propositional satisfiability (SAT) solvers~\cite{biere2009handbook},  satisfiability modulo theory (SMT) solvers~\cite{barrett2018satisfiability}, program synthesizers~\cite{jha2010oracle}, and proof assistants~\cite{nipkow2002isabelle}. While inductive learning techniques have been typically used for tasks such as perception  requiring decisions over very high-dimensional inputs~\cite{jha2018detecting}, the deductive AI methods have been successfully used to solve complex tasks such as automated planning, program reverse engineering, code repair, and protocol design. The remarkable success of LLMs has made it feasible to use them even for these complex tasks, and several recent results have demonstrated that LLMs can be used for program synthesis and planning~\cite{chen2021evaluating,song2022llm}. More generally, the LLMs exhibit limited symbolic reasoning capability making them applicable to problems that traditionally required symbolic AI reasoning methods such as SAT and SMT solving. 

Despite the reasoning capability of LLMs being extremely limited compared to symbolic reasoning methods, they overcome one key hindrance that has limited the widespread use of formal methods and automated reasoning: the need for deep expertise in formalizing and specifying problems to these tools. LLMs only require natural language specification of the problem and have been shown to be capable of automated formalization. Further, the output of the LLMs can be accompanied with easy to understand natural language explanations. This ease of interaction with LLMs as reasoning engines can enable non-expert users to pose planning or program synthesis prompts to them and obtain a solution that is easy to understand. 
However, LLMs are prone to generating factually incorrect or contextually inappropriate responses, also referred to as hallucinations~\cite{maynez2020faithfulness,roller2020recipes,cao2022hallucinated,ji2022survey,jha2021protein,kaur2022idecode}. The lossy encoding of knowledge in LLMs and the memory distortion that accompanies knowledge generalization naturally lead to inaccurate retrieval of knowledge even from training data. Thus, simply scaling data and models is not sufficient to resolve this problem.

\section{Related Work}
The use of large language models for tasks such as planning and code generation has been an area of active research in recent years, and numerous techniques have been developed to exploit the capabilities of these models in these domains.
Large language models have been used to generate plans for complex tasks~\cite{song2022llm,huang2022language} 
and for code generation~\cite{chen2021evaluating,wang2021codet5,nijkamp2022codegen}.
Several efforts to improve the performance of large language models in code generation and planning have been recently made through the use of specialized training techniques such as a combination of contrastive learning and fine-tuning~\cite{liu2023contrabert}, incorporation of external feedback in the form of success detection and human interaction through internal monologue~\cite{huang2022inner}, and decomposition of plans to sub-plans using LLMs followed by mapping to admissible actions using demonstrations~\cite{huang2022language}. 
Despite these advancements, several challenges remain in the application of large language models in planning and code generation. Notably, the model's ability to handle edge cases, produce consistent outputs, and understand the precise semantics of the generated code or plan is very limited~\cite{valmeekam2023planning,valmeekam2022large}. 
The lack of world knowledge and contextual understanding is another limitation of large language models in planning. These models struggle to consider real-world constraints and commonsense reasoning in their planning process, which often leads to the generation of unrealistic or unfeasible plans~\cite{marcus2020next}.
Further, the opacity of the decision-making process in large language models is a crucial limitation for the use of the LLM-produced programs or plans. 

In order to address this problem and make the LLM outputs more consistent and robust, and consequently less prone to errors and hallucinations, fine-tuning has been proposed to adapt to specific tasks or domains~\cite{lee2019mixout}. But for many domains, it is not obvious what a sufficient size of the domain-specific dataset needs to be used for fine-tuning, and continuously monitoring and fine-tuning LLMs is not practical. Failure of fine-tuning to improve LLM performance has been reported in literature~\cite{bommarito2022gpt}. 
Further, fine-tuning~\cite{wang2022preserving} adversely impacts the model's fluency, conversational capability, and in-context learning ability, which is critical to its response to prompts. 

Another approach is to connect LLMs to knowledge graphs~\cite{wang2017knowledge}, which represent knowledge as a graph of interconnected entities and relationships. Knowledge graphs can encode a wide range of structured and unstructured knowledge, including facts, concepts, and relationships.
Methods have been developed to infuse structured knowledge into LLMs by %
training models on factual triples of knowledge graphs (KGs), and such  models pre-trained on knowledge graphs have been shown to outperform baselines~\cite{moiseev2022skill}. But this requires a well-curated and complete knowledge base, building which is a time-consuming and expensive endeavor. Maintaining these knowledge bases with consistent and up-to-date information over time is also challenging. 

Yet another possibility is to encode 
external relevant knowledge into a key-value memory that exploits the fast maximum inner product search for memory querying. These memory slots can then be integrated with language models~\cite{wu2022efficient} for relatively smaller models, such as T5. Such a memory augmentation has been shown to improve the performance of the deep learning model on knowledge-intensive tasks. More recently, recurrent memory transformer (RMT) has been shown to be computationally efficient for large prompts with a million or more tokens~\cite{bulatov2023scaling}.

But for tasks such as programming or plan generation, the space of possible queries  is very large and can have many syntactic variations. Some of these simple syntactic variations could be superfluous with respect to the actual task, and not be relevant to the planning problem. 
It has been recently shown that LLMs are very sensitive to such irrelevant variations~\cite{jha2022responsible}.  
Hence, improving accuracy via expensive fine-tuning or explicit curation of knowledge graphs in such a context would be unrealistic as it would require not just creating variations relevant to the domain but also considering changes irrelevant to the core search problem. We also aim to avoid making significant changes to the deep learning model's architecture, such as memory augmentation, which can cause a significant decline in the scalability and accuracy of these models.

Our approach to generating trustworthy formal artifacts using LLMs is based on formal synthesis~\cite{jha2011towards,jha2017theory,jha2010oracle,solar2006combinatorial}. %
Formal synthesis generates a program, a plan, or a protocol satisfying a high-level formal specification.
In particular, we use the paradigm of counterexample-guided inductive synthesis~\cite{solar2006combinatorial} - an instance of oracle-guided inductive synthesis~\cite{jha2017theory}, wherein the oracle is a verifier, and the feedback provided to the learning engine is in the form of counterexamples.
These formal approaches to synthesis 
blend induction and deduction in the sense that
even as they generalize from examples, deductive procedures are used in the process of generalization.  
Formal inductive synthesis and machine learning fields have the same high-level goal: to develop algorithmic techniques for synthesizing a concept from observations. However, there are also important differences in the problem formulations
and techniques used in both fields. First, the concept classes in machine learning tend to use optimization-friendly representations such as deep neural networks, convex polytopes, or half-spaces. In the case of formal synthesis, the target of learning includes general programs, protocols, or automata. Secondly, the learning process in formal synthesis often incorporates combinatorial search methods such as satisfiability solving~\cite{biere2009handbook}, and hence, are much less scalable compared to purely learning techniques. But these approaches provide formal guarantees on the produced artifacts, which is lacking in learning-based approaches. This paper combines counterexample-guided inductive synthesis with LLMs for the scalable yet trusted synthesis of plans and programs.

\begin{figure}[t]
\centering
  \centering  \includegraphics[width=0.99\linewidth]{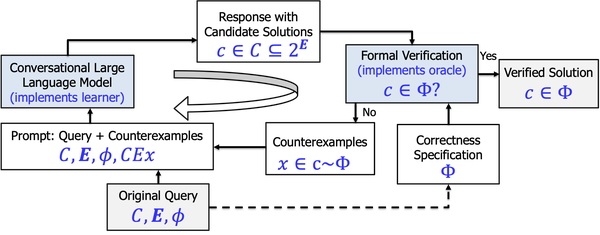}
\caption{The technical approach in this paper wraps the generative capability of the large language models in a formal synthesis loop inspired from counterexample-guided inductive synthesis~\cite{solar2006combinatorial} and more generally, oracle-guided inductive synthesis~\cite{jha2017theory}. The learning oracle in our approach is the conversation large language model, and the verification engine serves as the oracle. In particular, we use SMT solvers~\cite{de2008z3} for verification. The initial query defines the concept class (plans over a given set of actions), the set of possible behaviors (state space and the evolution of the environment state), and the specification. The verifier complements this initial query by adding counterexamples --- behaviors of the produced incorrect plans by the LLM that are not consistent with the specification ($\sim$ denotes symmetric set difference). The final output is the candidate produced by the LLM that is formally verified to be consistent with the given specification.}
\label{fig:exp_resnet}
\end{figure}

\section{Technical Approach}

We begin by defining some basic terms and notations motivated by the literature on formal synthesis~\cite{jha2017theory}. Following standard terminology in
the machine learning theory community~\cite{angluin1988queries}, we define 
a concept $\concept$ as a set of examples drawn from a domain of examples
$\domain$. In other words, $\concept \subseteq \domain$. 
For example, when learning to produce plans, the $\domain$ could be the sequence of all possible state evolutions of the world, and the plan could be the subset of evolutions that correspond to the correct action sequence. 
The set of all possible concepts is termed the {\em concept class},
denoted by $\class$. Thus, $\class \subseteq \powset{\domain}$.
Depending on the application domain, $\domain$ can be finite or
infinite. The concept class $\class$ can also be finite or
infinite. Note that it is possible to place (syntactic) restrictions on concepts so that $\class$ is finite even when $\domain$ is infinite. Unlike machine learning, one distinguishing characteristic of our problem formulation is the presence of an explicit formal
specification. This is crucial for an unambiguous definition of correct output. 

While the complete specification is infeasible, many domains, such as planning and program synthesis, admit properties that characterize the target plan or program requirements. Thus, a specification $\Spec$ 
can be thought to represent a set of ``correct'' concepts, i.e., $\Spec \subseteq
\class \subseteq \powset{\domain}$. 
Any example $\ex \in \domain$ such that there is a concept $c \in
\Spec$ where $\ex \in c$ is called a positive example or witness. 
Likewise, an example $\ex$ that is not contained
in any $c \in \Spec$ is a  negative example or counterexample. 
Given a concept $\target \in \class$ we say that $\target$ satisfies
$\Spec$ iff $\target \in \Spec$.
If we have a complete specification, it means that $\Spec$ is a
singleton set comprising only a single allowed concept.
In general, $\Spec$ is likely to be a partial specification
that allows for multiple correct concepts.

Since we use LLMs for learning, the subsets of examples can be provided as part of the prompts and used for in-context learning by the LLMs. Using only the witnesses is common in machine learning, but the presence of deductive reasoning engines and formal verification methods can enable a richer set of inputs, such as counterexamples obtained from verifying whether the artifact produced from the LLM satisfies the given specification. In general, we can model the availability of this richer set of inputs that can access the specification  
 $\Spec$ as an {\em oracle interface}. 
An {\em oracle interface} $\orint$ is a subset of $\defquery \times \defresponse$ where
$\defquery$ is a set of query types, 
$\defresponse$ is a corresponding set of response types, and
$\orint$ defines which pairs of query and response types are semantically well-formed.
A simple instance of an oracle interface is one with a single query type that
returns positive examples from $\domain$. Another instance is an oracle that has access to a verification engine and can return counterexamples. 
Implementations of the oracle interface can be nondeterministic algorithms
that exhibit nondeterministic choices in the stream of queries and responses.
Consider a {\em concept class} $\class$, a domain of examples
$\domain$, a specification $\Spec$, and an oracle interface $\orint$. 
Our goal is to find
a {\em target concept} $\target \in \class$ that
satisfies $\Spec$,
given only $\orint$ and $\class$. 
In other words, 
 $\domain$ and $\Spec$ can be accessed  only through $\orint$.

We restrict the oracle to be a verifier that can produce counterexamples, and thus, our approach becomes an instance of counterexample-guided inductive synthesis (CEGIS).
It was originally proposed as an algorithmic method for program
synthesis where the specification is given as a reference program and
the concept class is defined using a partial program, also referred to
as a ``sketch''~\cite{solar2006combinatorial}. 
In CEGIS, the learner (synthesizer) interacts with a ``verifier" that can take in a candidate plan or a program and
a specification, and 
try to find a counterexample showing that the candidate program
does not satisfy the specification. 
In CEGIS, the learner is typically implemented on top of a
general-purpose decision procedure such as a satisfiability solver, satisfiability modulo theory solver,
or model checker. The oracle (verifier) is also implemented similarly. 
In contrast, we use the LLMs as the learner in our work and a formal verification engine as the oracle. 

Thus, we adopt the CEGIS architecture and locate the LLM as a scalable learner within it. This enables us to scale synthesis and, at the same time, produce plans or programs that are formally verified against the given specifications. 
This proposed approach extends our recent work-in-progress report on improving the trust of LLMs~\cite{jha-icaa23}, and exploits the following characteristics of the LLMs and the satisfiability solvers as verification engines.
First, the LLMs can be prompted to generate well-structured outputs that can be easily parsed by a formal verification engine. The inclusion of a large corpus of code in their training data is perhaps responsible for this characteristic. We successfully use this to ensure the output of LLMs can be ingested by the verifier. We also use LLMs to derive the queries to the verifier, automatically checking for the correctness specification.  Second, the formal verifiers are very fast at checking the correctness of a single solution for a task such as planning, as compared to solving the planning problem using combinatorial search and finding a valid plan. Formal verification techniques can also be used to localize the part of the solution that violates the specification and to provide informative feedback to the LLMs that rule out not just a single incorrect solution but a whole class of incorrect solutions. 
A large body of work on verification, fault isolation, and explanation generation can be leveraged using this architecture. This would expand the query types and response types ($\defquery, \defresponse$) in the formal synthesis framework. Further, LLMs demonstrate remarkable in-context learning, and adding counterexamples and explanations to the prompt steers them away from incorrect responses and eventually drives them to a correct solution that is accepted by the verifier. 
The context length used to train LLMs has been rapidly growing. Models such as GPT-4 can take a context length of 32K tokens, while other generative models with 100K tokens have also been developed. This larger context length enables LLMs to stay consistent with provided prompts. This increasing context length enables being able to support more number of counterexamples in the  prompt. Thus, the current direction of improvement in LLMs supports more number of iterations of the presented architecture. Finally, the use of formal methods for tasks such as planning requires tedious modeling into existing solvers and verifiers. The responses of these formal engines also need to be translated back into more easily interpretable forms to be used by engineers and developers who are not experts in formal methods. The use of LLM in the above architecture also serves as a human-friendly front end and, thus, partially alleviates the challenge of making formal synthesis engines more accessible to non-expert users. 

In the rest of the paper, we evaluate the proposed approach using automated planning as the synthesis domain and using SMT solver~\cite{de2008z3} as the verification engine. We first describe a case study in detail and then present the quantitative results.

\section{Experiments}

We use block-world planning~\cite{gupta1992complexity} traditionally used in the traditional planning literature for our experiments. 
This standard benchmark allows us to test scalability and accuracy across different problem sizes and using different LLMs. 
We  specify the high-level predicates that capture the state of the world, such as \texttt{BlockOn, OnTable,} and \texttt{Holding}, and their evolution. We  describe the permitted operations/actions and the initial state. We provide a final state and ask the LLM to produce a valid plan. The LLMs take a few iterations with the verifier to produce the correct result. We consider responses where the verifier appends the incorrect sequences to the prompt only stating that these sequences are not valid or provides a richer response as a prefix of the invalid plan such that any completion would be incorrect. This iteration continues till the response of the LLM is verified to be correct. 

The final response from the LLM provides a human-readable justification in addition to the final plan. 
We use one example of a block-world problem to demonstrate our proposed approach before presenting quantitative results that compare different LLMs, such as GPT4, GPT3.5 Turbo, Davinci, Curie, Babbage, and Ada, and also consider planning problems of varying scale. 

\noindent\textbf{Illustrative Example} 
We consider the block world planning problem.
The natural language specification of the plan in the BLOCKWORLD problem is also translated into formal (first-order) constraints using the LLM. We provide the details for that step in the appendix
and focus here on the iterative refinement of the prompt to steer LLM toward a correct response using the formal verifier as the guiding oracle. We provide two examples in our prompt to assist the LLMs  to inductively reason about a new problem and respond with its plan. 

\noindent One of the two examples is listed below.

\lstset{
backgroundcolor=\color{gray!10},   basicstyle=\ttfamily \fontsize{7pt}{7.5pt}\selectfont,
    breaklines=true,
    keywordstyle=[1]\color{blue},
    keywordstyle=[2]\color{blue},
    keywordstyle=[3]\color{blue},
    keywordstyle=[4]\color{blue},
    keywordstyle=[5]\color{blue},
    stringstyle=\color{red},
commentstyle=\color{green!60!black},
    frame=single,
    language={},
    tabsize=4,
    numbers=left,
numberstyle=\tiny\color{gray},
columns=flexible,
xleftmargin=.04\textwidth, xrightmargin=.04\textwidth,
captionpos=b
}

\lstdefinestyle{customblock}{
    morekeywords=[1]{define},
    morekeywords=[2]{objects},
    morekeywords=[3]{init},
    morekeywords=[4]{goal},
    morekeywords=[5]{domain}
}

\begin{lstlisting}[style=customblock,numbers=none]
Given the block world problem OLDPROB1:
(define (problem BW-sample-0)
(:domain blocksworld-4ops)
(:objects b1 b2 b3 b4 b5 b6 )

(:init
(arm-empty)
(on b1 b2)
(on b2 b3)
(on b3 b4)
(on b4 b5)
(on b5 b6)
(on-table b6)
(clear b1)
(clear b2)
(clear b3)
(clear b4)
(clear b5)
(clear b6)
)

(:goal
(and
(on b6 b5)
(on b5 b4)
(on b4 b3)
(on b3 b2)
(on b2 b1)
(on-table b1)
)))
\end{lstlisting}

\noindent We first list the problem, as shown above, and then a plan for the planning problem. 

\begin{lstlisting}[style=customblock,numbers=none]
The solution for the problem OLDPROB1 is:
START-PLAN
1. unstack b1 b2
2. put-down b1
3. unstack b2 b3
4. put-down b2
5. unstack b3 b4
6. put-down b3
7. unstack b4 b5
8. put-down b4
9. unstack b5 b6
10. put-down b5
11. pick-up b6
12. stack b6 b5
13. pick-up b5
14. stack b5 b4
15. pick-up b4
16. stack b4 b3
17. pick-up b3
18. stack b3 b2
19. pick-up b2
20. stack b2 b1
END-PLAN
\end{lstlisting}

\noindent  Now, given a new block world problem NEWPROB, the LLM may produce an incorrect response:
\begin{lstlisting}[style=customblock,numbers=none]
(define (problem BW-rand-6)
(:domain blocksworld-4ops)
(:objects b1 b2 b3 b4 b5 b6 )
(:init
(arm-empty)
(on b1 b6)
(on b2 b3)
(on-table b3)
(on b4 b1)
(on-table b5)
(on-table b6)
(clear b2)
(clear b4)
(clear b5)
)
(:goal
(and
(on b1 b2)
(on b5 b3)
(on b6 b4))
)
) 
\end{lstlisting}

\noindent  The incorrect solution from the problem NEWPROB is:
\begin{lstlisting}[style=customblock,numbers=none]
START-PLAN
1. unstack b4 b1
2. put-down b4
3. unstack b1 b6
4. put-down b1
5. pick-up b2
6. stack b2 b3
7. pick-up b1
8. stack b1 b2
9. pick-up b5
10. stack b5 b3
11. pick-up b6
12. stack b6 b4
END-PLAN
\end{lstlisting}

\noindent The SMT solver identifies the following as an infeasible subset of the plan and generates a new prompt including the statement that ``Any plan with the following prefix is not correct'':
\begin{lstlisting}[style=customblock,numbers=none]
START-PLAN
1. unstack b4 b1
2. put-down b4
3. unstack b1 b6
4. put-down b1
5. pick-up b2
\end{lstlisting}
Using this information, the LLM produces a new response that is verified to be correct by the decision procedure.
\begin{lstlisting}[style=customblock,numbers=none]
START-PLAN
1. unstack b4 b1
2. put-down b4
3. unstack b1 b6
4. put-down b1
5. unstack b2 b3
6. put-down b2
7. pick-up b5
8. stack b5 b3
9. pick-up b6
10. stack b6 b4
11. pick-up b1
12. stack b1 b2
END-PLAN
\end{lstlisting}

\begin{figure}[h]
    \centering
\resizebox{0.45\textwidth}{!}{%
\begin{tikzpicture}
\begin{axis}[
    ybar,
    enlargelimits=0.15,
    legend style={at={(0.5,-0.2)},
      anchor=north,legend columns=-1},
    ylabel={Number of Successful Plans},
    symbolic x coords={GPT-4, GPT-3.5 Turbo, Davinci, Curie, Babbage, Ada},
    xtick=data,
    nodes near coords,
    nodes near coords align={vertical},
    x tick label style={rotate=45,anchor=east},
    ]
\addplot coordinates {(GPT-4,19) (GPT-3.5 Turbo,12) (Davinci,2) (Curie,2) (Babbage,2) (Ada,0)};
\end{axis}
\end{tikzpicture}
}
    \caption{The performance of different GPT models on problems with 3 blocks.}
    \label{fig:gptall3blocks}
\end{figure}
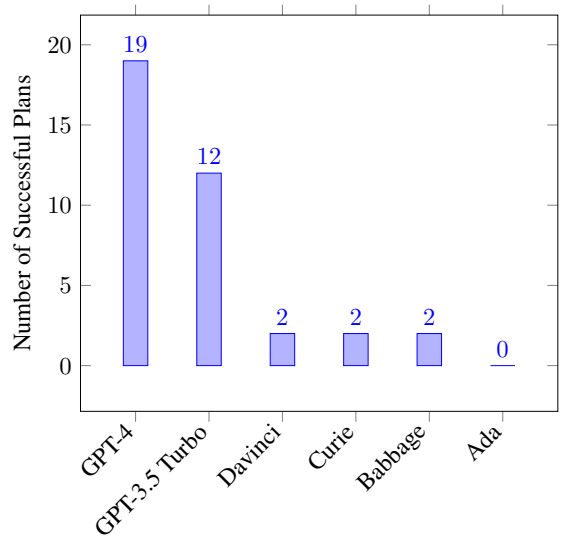

\noindent \textbf{Quantitative Results}
We analyze the performance of different GPT on 20 random problems with 3 blocks in Fig.~\ref{fig:gptall3blocks}.

\begin{figure}[h]
\centering
\resizebox{0.45\textwidth}{!}{%
\begin{tikzpicture}
\begin{axis}[
    ybar,
    bar width=0.5cm,
    width=0.475\textwidth,
    height=6cm,
    enlarge x limits=0.15,
    legend style={at={(0.5,-0.15)}, anchor=north,legend columns=-1},
    ylabel={Number of Correct Solutions},
    symbolic x coords={3 Blocks,4 Blocks,5 Blocks,6 Blocks},
    xtick=data,
    nodes near coords,
    nodes near coords align={vertical},
    ymin=0,
    x tick label style={rotate=45,anchor=east},
    ]
\addplot[fill=blue!50] coordinates {(3 Blocks,12) (4 Blocks,4) (5 Blocks,6) (6 Blocks,1)};
\end{axis}
\end{tikzpicture}
}
    \caption{The performance of GPT-3.5 on problems with multiple blocks.}
    \label{fig:gpt3allblocks}
\end{figure}
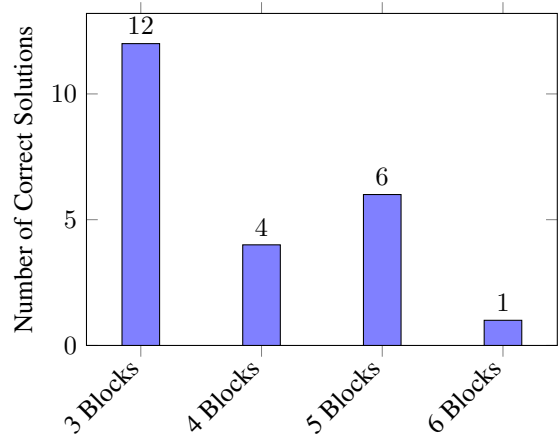

GPT-4 is the most successful model and we analyze its performance with with 3, 4, 5 and 6 blocks. In Fig.~\ref{fig:gpt4allblocks}, we observe that it is only able to solve 2 out of the 20 problems with 10 blocks even though it could solve 19 out of the 20 problem with 3 blocks.  We also study the performance of the second best model, that is, GPT 3.5 with a different number of blocks, and the results are shown in Fig.~\ref{fig:gpt3allblocks}.

\begin{figure}[h]
\centering
\resizebox{0.4\textwidth}{!}{%
\begin{tikzpicture}
\begin{axis}[
    ybar,
    bar width=0.5cm,
    width=0.475\textwidth,
    height=6cm,
    enlarge x limits=0.15,
    legend style={at={(0.5,-0.15)}, anchor=north,legend columns=-1},
    ylabel={Number of Correct Solutions},
    symbolic x coords={3 Blocks,4 Blocks,5 Blocks,6 Blocks, 10 Blocks},
    xtick=data,
    nodes near coords,
    nodes near coords align={vertical},
    ymin=0,
    x tick label style={rotate=45,anchor=east},
    ]
\addplot[fill=blue!50] coordinates {(3 Blocks,19) (4 Blocks,11) (5 Blocks,11) (6 Blocks,9) (10 Blocks, 2)};
\end{axis}
\end{tikzpicture}
}
    \caption{The performance of GPT-4 on problems with multiple blocks.}
    \label{fig:gpt4allblocks}
\end{figure}
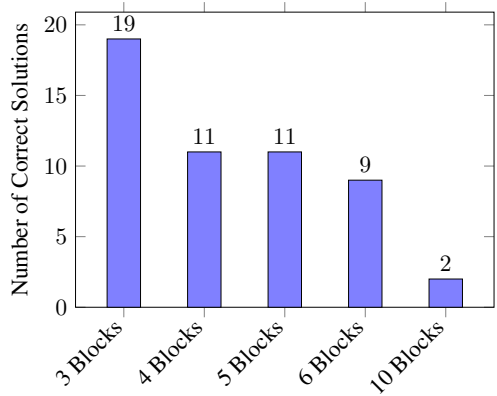

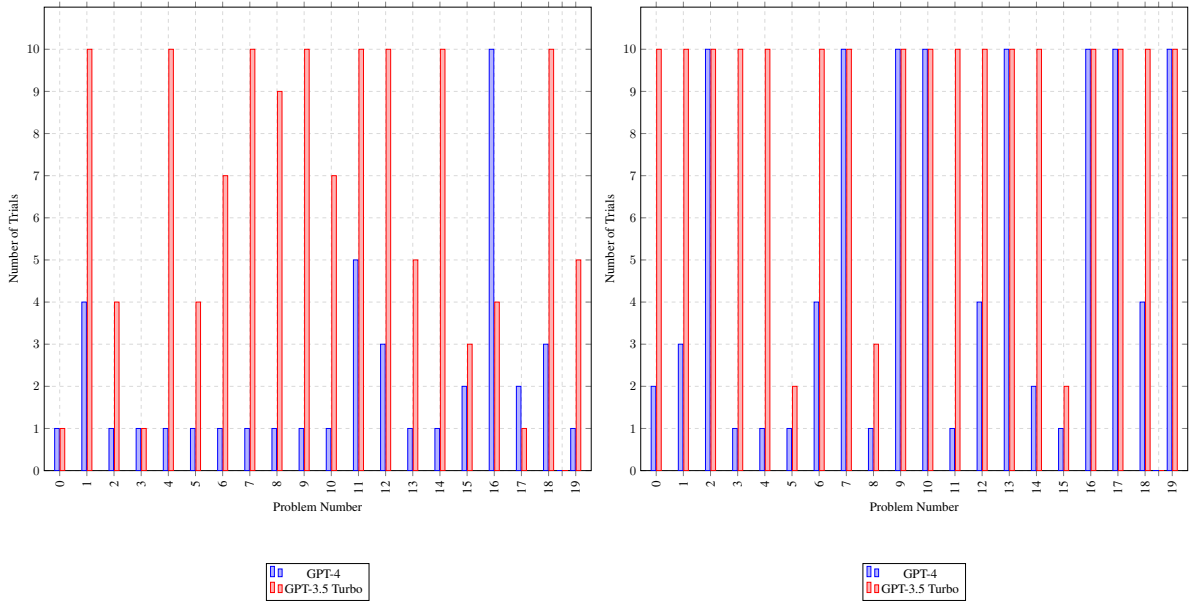
\begin{figure}[h]
\begin{center}
\resizebox{0.475\textwidth}{!}{%
\begin{tikzpicture}
\begin{axis}[
    width=\linewidth, %
    grid=major, %
    grid style={dashed,gray!30}, %
    xlabel=Problem Number, %
    ylabel=Number of Trials,
    legend style={at={(0.5,-0.2)},anchor=north},
    x tick label style={rotate=90,anchor=east},
    symbolic x coords={0,,1,,2,,3,,4,,5,,6,,7,,8,,9,,10,,11,,12,,13,,14,,15,,16,,17,,18,,19},
    xtick=data,
    ybar=0.025cm, %
    bar width=0.125cm, %
    ymin=0,
    ymax=11,
    ytick={0,1,2,3,4,5,6,7,8,9,10},
    enlarge x limits=0.03, %
]
\addplot coordinates {
    (0,1)
    (,0)
    (1,4)
    (,0)
    (2,1)
    (,0)
    (3,1)
    (,0)
    (4,1)
    (,0)
    (5,1)
    (,0)
    (6,1)
    (,0)
    (7,1)
    (,0)
    (8,1)
    (,0)
    (9,1)
    (,0)
    (10,1)
    (,0)
    (11,5)
    (,0)
    (12,3)
    (,0)
    (13,1)
    (,0)
    (14,1)
    (,0)
    (15,2)
    (,0)
    (16,10)
    (,0)
    (17,2)
    (,0)
    (18,3)
    (,0)
    (19,1)
};
\addplot coordinates {
    (,0)
    (0,1)
    (,0)
    (1,10)
    (,0)
    (2,4)
    (,0)
    (3,1)
    (,0)
    (4,10)
    (,0)
    (5,4)
    (,0)
    (6,7)
    (,0)
    (7,10)
    (,0)
    (8,9)
    (,0)
    (9,10)
    (,0)
    (10,7)
    (,0)
    (11,10)
    (,0)
    (12,10)
    (,0)
    (13,5)
    (,0)
    (14,10)
    (,0)
    (15,3)
    (,0)
    (16,4)
    (,0)
    (17,1)
    (,0)
    (18,10)
    (,0)
    (19,5)
};
\legend{GPT-4, GPT-3.5 Turbo}
\end{axis}
\end{tikzpicture}
}
 \resizebox{0.475\textwidth}{!}{%
\begin{tikzpicture}
\begin{axis}[
    width=\linewidth, %
    grid=major, %
    grid style={dashed,gray!30}, %
    xlabel=Problem Number, %
    ylabel=Number of Trials,
    legend style={at={(0.5,-0.2)},anchor=north},
    x tick label style={rotate=90,anchor=east},
    symbolic x coords={0,,1,,2,,3,,4,,5,,6,,7,,8,,9,,10,,11,,12,,13,,14,,15,,16,,17,,18,,19},
    xtick=data,
    ybar=0.025cm, %
    bar width=0.125cm, %
    ymin=0,
    ymax=11,
    ytick={0,1,2,3,4,5,6,7,8,9,10},
    enlarge x limits=0.03, %
]
\addplot coordinates {
    (0,2)
    (,0)
    (1,3)
    (,0)
    (2,10)
    (,0)
    (3,1)
    (,0)
    (4,1)
    (,0)
    (5,1)
    (,0)
    (6,4)
    (,0)
    (7,10)
    (,0)
    (8,1)
    (,0)
    (9,10)
    (,0)
    (10,10)
    (,0)
    (11,1)
    (,0)
    (12,4)
    (,0)
    (13,10)
    (,0)
    (14,2)
    (,0)
    (15,1)
    (,0)
    (16,10)
    (,0)
    (17,10)
    (,0)
    (18,4)
    (,0)
    (19,10)
};
\addplot coordinates {
    (,0)
    (0,10)
    (,0)
    (1,10)
    (,0)
    (2,10)
    (,0)
    (3,10)
    (,0)
    (4,10)
    (,0)
    (5,2)
    (,0)
    (6,10)
    (,0)
    (7,10)
    (,0)
    (8,3)
    (,0)
    (9,10)
    (,0)
    (10,10)
    (,0)
    (11,10)
    (,0)
    (12,10)
    (,0)
    (13,10)
    (,0)
    (14,10)
    (,0)
    (15,2)
    (,0)
    (16,10)
    (,0)
    (17,10)
    (,0)
    (18,10)
    (,0)
    (19,10)
};
\legend{GPT-4, GPT-3.5 Turbo}
\end{axis}
\end{tikzpicture}
}
\end{center}
\caption{Number of iterations required by GPT-4 and GPT-3.5 Turbo.}
\label{fig:iterations}
\end{figure}

We study how LLMs are improved by adding verification in the loop and report the number of iterations taken by our approach to produce the correct result. In Fig.~\ref{fig:iterations}, we analyze the number of iterations of counterexamples need by GPT-4 and GPT-3.5 on 20 random problems involving 3 and 4 blocks.

\section{Conclusions}
We present a novel approach to addressing the  challenge of the trustworthy generation of formal artifacts such as plans and programs from LLMs. Our technique uses the paradigm of counterexample-guided inductive synthesis wherein the learner is implemented using the LLM and the verifier uses SMT solver. 
Our approach can  be viewed as an adversarial variant~\cite{jha2019attribution} of the in-context learning approach, wherein we use formal verification to detect incorrect responses and include the generated counterexamples as a part of the prompt in the dialog with the LLM. 
The initial experiments reported over the planning task in this  paper are encouraging and indicate that the proposed combination of LLMs and formal verifiers can be an effective approach to using LLMs in applications where the generated artifact must be verified.

\bibliographystyle{alpha}
\bibliography{milcom}

\newpage 

\appendix

\section{Experimental Details}

The natural language specification of the plan in the BLOCKWORLD problem is also translated into formal ( first order ) constraints using the LLM. For this, we specify the key predicates and then use the prompt ``Translate the following into first-order logic using the formatting of Z3 Python API and the following state information:'' to enable the translation and obtain the corresponding Z3 API code that we then deploy in our deductive reasoning engine. The state information given to the LLM is as follows:

\lstset{
    backgroundcolor=\color{white},
    basicstyle=\ttfamily, %
    breaklines=true,
    keywordstyle=[1]\color{blue},
    keywordstyle=[2]\color{blue},
    keywordstyle=[3]\color{blue},
    keywordstyle=[4]\color{purple},
    keywordstyle=[5]\color{purple},
    stringstyle=\color{red},
    commentstyle=\color{green!60!black},
    frame=single,
    language=Python,
    tabsize=4,
    numbers=left,
    numberstyle=\tiny\color{gray},
columns=flexible,
xleftmargin=.04\textwidth, xrightmargin=.04\textwidth,
    captionpos=b
}

\lstdefinestyle{custompython}{
    morekeywords=[1]{class},
    morekeywords=[2]{def},
    morekeywords=[3]{Function},
    morekeywords=[4]{IntSort},
    morekeywords=[5]{BoolSort}
}

\begin{lstlisting}[style=custompython]
class State:
    def __init__(self, name):
        self.table = Function(f'{name}_table', IntSort(), BoolSort())
        self.hand = Function(f'{name}_hand', IntSort(), BoolSort())
        self.stacked = Function(f'{name}_stacked', IntSort(), IntSort(), BoolSort())
        self.clear = Function(f'{name}_clear', IntSort(), BoolSort())
        self.handsfree = Function(f'{name}_handsfree', IntSort(), BoolSort())
\end{lstlisting}

The informal specification for the Pick-up operation in the planning problem is given by: 

\begin{center}
\begin{tcolorbox}[colback=gray!10, colframe=gray!50, width=0.9\textwidth]
\footnotesize
The block must be clear and on the table in state s1. The block must not be in the hand in state s1. The block must be in the hand in state s2. The block must not be on the table in state s2 and must not be clear. 
For all other blocks x, the state of the hand, table, and clear should remain the same between s1 and s2.
For all blocks x, y, the state of the stacked should remain the same between s1 and s2.
\end{tcolorbox}
\end{center}

\noindent The corresponding produced formal constraints by the LLM are: 

\begin{center}
\begin{tcblisting}{
width=0.9\textwidth,
    listing engine=minted,
    colback=black!99!white,
    colframe=black!10!black,
    listing only,
    minted style=native,
    minted language=python,
minted options={baselinestretch=1, %
breaklines,
xleftmargin=.001\textwidth, 
xrightmargin=.001\textwidth,}}
# Constraint 1: The block must be clear and on the table in state s1
constraint1 = And(s1.table(the_block), s1.clear(the_block))
# Constraint 2: The block must not be in the hand in state s1
constraint2 = Not(s1.hand(the_block))
# Constraint 3: The block must be in the hand in state s2
constraint3 = s2.hand(the_block)
# Constraint 4: The block must not be on the table in state s2 and must not be clear
constraint4 = And(Not(s2.table(the_block)), Not(s2.clear(the_block)))
# Constraint 5: For all other blocks x, the state of the hand, table, and clear should remain the same between s1 and s2
x = Int('x')
constraint5 = ForAll([x], And(s1.table(x) == s2.table(x), s1.hand(x) == s2.hand(x), s1.clear(x) == s2.clear(x)), patterns=[x != the_block])
# Constraint 6: For all blocks x, y, the state of the stacked should remain the same between s1 and s2
x = Int('x')
y = Int('y')
constraint6 = ForAll([x, y], s1.stacked(x, y) == s2.stacked(x, y))
# Combine all constraints
constraints = And(constraint1, constraint2, constraint3, constraint4, constraint5, constraint6)
\end{tcblisting}
\end{center}

We query the LLM to translate the following constraint into first order logic for the Put-down operation in the planning problem:
\begin{center}
\begin{tcolorbox}[colback=gray!10, colframe=gray!50, width=0.9\textwidth]
\footnotesize
The block must not be in the hand in state s2. The block must be on the table in state s2 and must be clear.
For all other blocks x, the state of the hand, table, and clear should remain the same between s1 and s2.
For all blocks x, y, the state of the stacked should remain the same between s1 and s2.
\end{tcolorbox}
\end{center}

Similarly, we use the following query to the LLM to translate the stack operation into first-order logic:

\begin{center}
\begin{tcolorbox}[colback=gray!10, colframe=gray!50, width=0.9\textwidth]
\footnotesize
 In all cases, the following should be true when executing the stack(block1, block2) operation. For all other blocks x other than these two blocks, the state of the hand, table, and clear should remain the same between the states s1 and s2. For all blocks x, y, where neither x is block1, nor y is block2, the state of being stacked should remain the same between s1 and s2.

There are two cases: either a block is stacked from the table or a block is stacked from the hand.

If the block1 is being stacked from the table, then the following is true: The block1 must be on the table in state s1. Neither block1 nor block2 should be stacked on each other in state s1. Both block1 and block2 must be clear, i.e., not having any block on them, in state s1. The agent's hand should be free in state s1. The block1 must not be in the agent's hand nor on the table in state s2. The block1 must be stacked on block2 in state s2. The block1 must be clear, while block2 must not be clear in state s2. The agent's hand should be free in state s2.

If the block1 is being stacked from the hand, then the following is true: The  block1 must be in the agent's hand in state s1. Neither block1 nor block2 should be stacked on each other in state s1. The block2 must be clear in state s1. The agent's hand should not be free in state s1. The block1 must not be in the agent's hand nor on the table in state s2. The block1 must be stacked on block2 in state s2. The block1 must be clear, while block2 must not be clear in state s2. The agent's hand should be free in state s2.

\end{tcolorbox}
\end{center}

The LLM produces the following first order logic formula using the flavored syntax of Z3 Python API. 

\begin{center}

\begin{tcblisting}{
width=0.9\textwidth,
    listing engine=minted,
    colback=black!99!white,
    colframe=black!10!black,
    listing only,
    minted style=native,
    minted language=python,
minted options={baselinestretch=1, fontsize=\fontsize{9.5pt}{9.5pt}\selectfont,
breaklines,
xleftmargin=.01\textwidth, 
xrightmargin=.01\textwidth,}}

constraints.append(ForAll(x, Implies(And(x != block1, x != block2), And(s1.hand(x) == s2.hand(x), s1.table(x) == s2.table(x), s1.clear(x) == s2.clear(x)))))

constraints.append(ForAll([x, y], Implies(And(x != block1, y != block2), s1.stacked(x, y) == s2.stacked(x, y))))

constraints.append(Or(
And(
# If the block1 is being stacked from the table
s1.table(block1),     # The block1 is on the table in state s1
Not(s1.stacked(block1, block2)),  # Neither block1 nor block2 are stacked on each other in state s1
s1.clear(block1),     # Block1 is clear in state s1
s1.clear(block2),     # Block2 is clear in state s1
s1.handsfree(),       # The agent's hand is free in state s1
Not(s2.table(block1)),  # Block1 is not on the table in state s2
Not(s2.hand(block1)),   # Block1 is not in the agent's hand in state s2
s2.stacked(block1, block2),  # Block1 is stacked on block2 in state s2
s2.clear(block1),     # Block1 is clear in state s2
Not(s2.clear(block2)),  # Block2 is not clear in state s2
s2.handsfree()        # The agent's hand is free in state s2
),
And(
# If the block1 is being stacked from the hand
s1.hand(block1),      # The block1 is in the agent's hand in state s1
Not(s1.stacked(block1, block2)),  # Neither block1 nor block2 are stacked on each other in state s1
s1.clear(block2),     # Block2 is clear in state s1
Not(s1.handsfree()),  # The agent's hand is not free in state s1
Not(s2.table(block1)),  # Block1 is not on the table in state s2
Not(s2.hand(block1)),   # Block1 is not in the agent's hand in state s2
s2.stacked(block1, block2),  # Block1 is stacked on block2 in state s2
s2.clear(block1),     # Block1 is clear in state s2
Not(s2.clear(block2)),  # Block2 is not clear in state s2
s2.handsfree()        # The agent's hand is free in state s2
)
))
\end{tcblisting}
\end{center}

\begin{center}
\begin{tcolorbox}[colback=gray!10, colframe=gray!50, width=0.9\textwidth]
In all cases, the following should be true when executing the unstack(block1, block2) operation. For all other blocks x other than these two blocks, the state of the hand, table, and clear should remain the same between the states s1 and s2. For all blocks x and y, where neither x is block1, nor y is block2, the state of being stacked should stay the same between s1 and s2.

There are two cases: either a block is unstacked to the table or a block is unstacked to the hand.

If the block1 is being unstacked from the block2 to the table, then the following must be true: The blocks block1 and block2 must be stacked in state s1, with block1 clear and the agent's hand free. In state s2, block1 should not be stacked on block2 but should be on the table. In state 2, both block1 and block2 should be clear, and the agent's hand should be free.

If the block1 is being unstacked from the block2 to the hand, then the following must be true: The blocks block1 and block2 must be stacked in state s1, with block1 clear and the agent's hand free. In state s2, block1 should not be stacked on block2 but should be in the agent's hand. The block2 should be clear, but block1 should not be clear, and the agent's hand should not be free.
\end{tcolorbox}
\end{center}

\subsection{Number of iterations with GPT-4 and GPT-3.5}

\subsubsection{GPT-4 and GPT-3.5 Turbo with 5 blocks}

 We analyze the number of iterations of computerexamples needed by GPT-4 and GPT-3.5 on 20 random problems with 5 blocks.

\begin{figure}[h]
\centering
\resizebox{0.75\textwidth}{!}{%
\begin{tikzpicture}
\begin{axis}[
    width=\linewidth, %
    grid=major, %
    grid style={dashed,gray!30}, %
    xlabel=Problem Number, %
    ylabel=Number of Trials,
    legend style={at={(0.5,-0.2)},anchor=north},
    x tick label style={rotate=90,anchor=east},
    symbolic x coords={0,,1,,2,,3,,4,,5,,6,,7,,8,,9,,10,,11,,12,,13,,14,,15,,16,,17,,18,,19},
    xtick=data,
    ybar=0.025cm, %
    bar width=0.125cm, %
    ymin=0,
    ymax=11,
    ytick={0,1,2,3,4,5,6,7,8,9,10},
    enlarge x limits=0.03, %
]
\addplot coordinates {
    (0,8)
    (,0)
    (1,10)
    (,0)
    (2,10)
    (,0)
    (3,10)
    (,0)
    (4,1)
    (,0)
    (5,10)
    (,0)
    (6,10)
    (,0)
    (7,1)
    (,0)
    (8,1)
    (,0)
    (9,8)
    (,0)
    (10,1)
    (,0)
    (11,2)
    (,0)
    (12,10)
    (,0)
    (13,8)
    (,0)
    (14,1)
    (,0)
    (15,1)
    (,0)
    (16,3)
    (,0)
    (17,10)
    (,0)
    (18,10)
    (,0)
    (19,10)
};
\addplot coordinates {
    (,0)
    (0,10)
    (,0)
    (1,10)
    (,0)
    (2,10)
    (,0)
    (3,10)
    (,0)
    (4,10)
    (,0)
    (5,10)
    (,0)
    (6,10)
    (,0)
    (7,10)
    (,0)
    (8,10)
    (,0)
    (9,4)
    (,0)
    (10,10)
    (,0)
    (11,5)
    (,0)
    (12,10)
    (,0)
    (13,10)
    (,0)
    (14,8)
    (,0)
    (15,2)
    (,0)
    (16,10)
    (,0)
    (17,6)
    (,0)
    (18,10)
    (,0)
    (19,3)
};
\legend{GPT-4, GPT-3.5 Turbo}
\end{axis}
\end{tikzpicture}
}
    \caption{Number of iterations required by GPT-4 and GPT-3.5 Turbo on problems with 5 blocks.}
    \label{fig:gpt4all5blocks}
\end{figure}
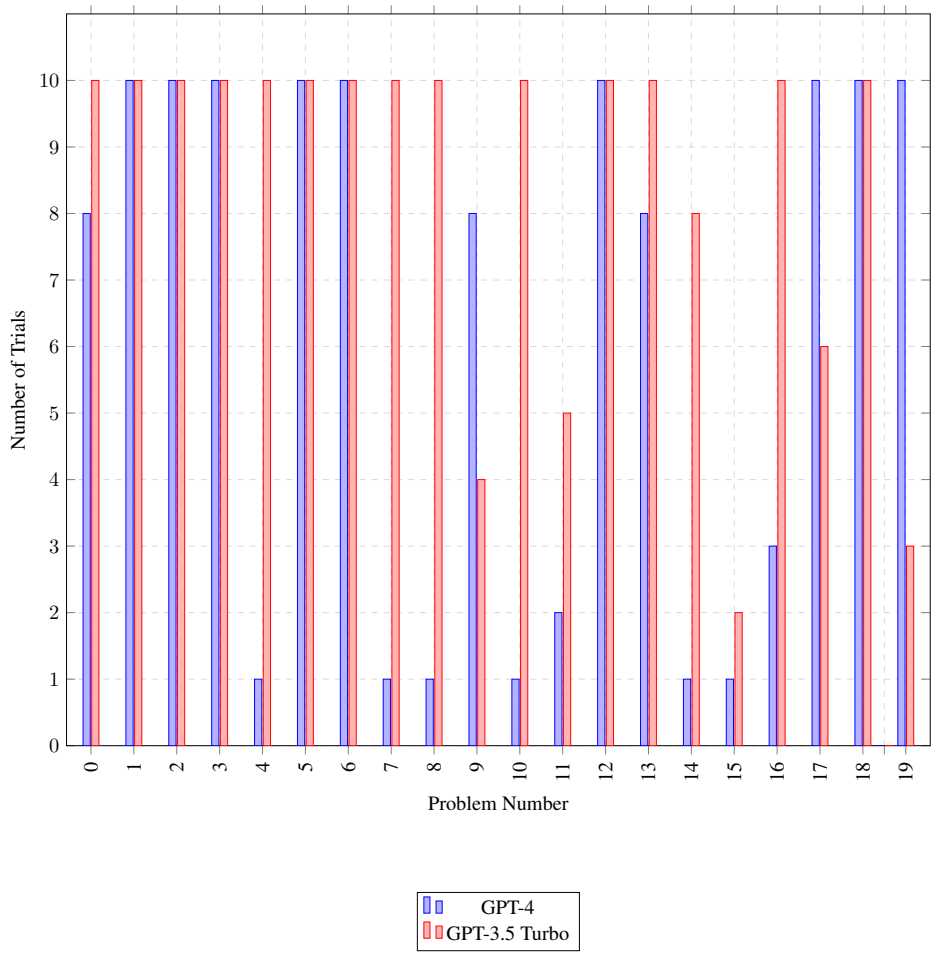

\subsubsection{GPT-4 and GPT-3.5 Turbo with 6 blocks}

 We analyze the number of iterations of computerexamples needed by GPT-4 and GPT-3.5 on 20 random problems with 6 blocks.

\begin{figure}[h]
\centering
\resizebox{0.75\textwidth}{!}{%
\begin{tikzpicture}
\begin{axis}[
    width=\linewidth, %
    grid=major, %
    grid style={dashed,gray!30}, %
    xlabel=Problem Number, %
    ylabel=Number of Trials,
    legend style={at={(0.5,-0.2)},anchor=north},
    x tick label style={rotate=90,anchor=east},
    symbolic x coords={0,,1,,2,,3,,4,,5,,6,,7,,8,,9,,10,,11,,12,,13,,14,,15,,16,,17,,18,,19},
    xtick=data,
    ybar=0.025cm, %
    bar width=0.125cm, %
    ymin=0,
    ymax=11,
    ytick={0,1,2,3,4,5,6,7,8,9,10},
    enlarge x limits=0.03, %
]
\addplot coordinates {
    (0,10)
    (,0)
    (1,5)
    (,0)
    (2,10)
    (,0)
    (3,5)
    (,0)
    (4,10)
    (,0)
    (5,10)
    (,0)
    (6,10)
    (,0)
    (7,1)
    (,0)
    (8,2)
    (,0)
    (9,10)
    (,0)
    (10,4)
    (,0)
    (11,4)
    (,0)
    (12,3)
    (,0)
    (13,10)
    (,0)
    (14,2)
    (,0)
    (15,10)
    (,0)
    (16,10)
    (,0)
    (17,10)
    (,0)
    (18,5)
    (,0)
    (19,10)
};
\addplot coordinates {
    (,0)
    (0,10)
    (,0)
    (1,10)
    (,0)
    (2,10)
    (,0)
    (3,10)
    (,0)
    (4,10)
    (,0)
    (5,10)
    (,0)
    (6,10)
    (,0)
    (7,10)
    (,0)
    (8,10)
    (,0)
    (9,10)
    (,0)
    (10,10)
    (,0)
    (11,3)
    (,0)
    (12,10)
    (,0)
    (13,10)
    (,0)
    (14,10)
    (,0)
    (15,10)
    (,0)
    (16,10)
    (,0)
    (17,10)
    (,0)
    (18,10)
    (,0)
    (19,10)
};
\legend{GPT-4, GPT-3.5 Turbo}
\end{axis}
\end{tikzpicture}
}
    \caption{Number of iterations required by GPT-4 and GPT-3.5 Turbo on problems with 6 blocks.}
    \label{fig:gpt6allblocks}
\end{figure}
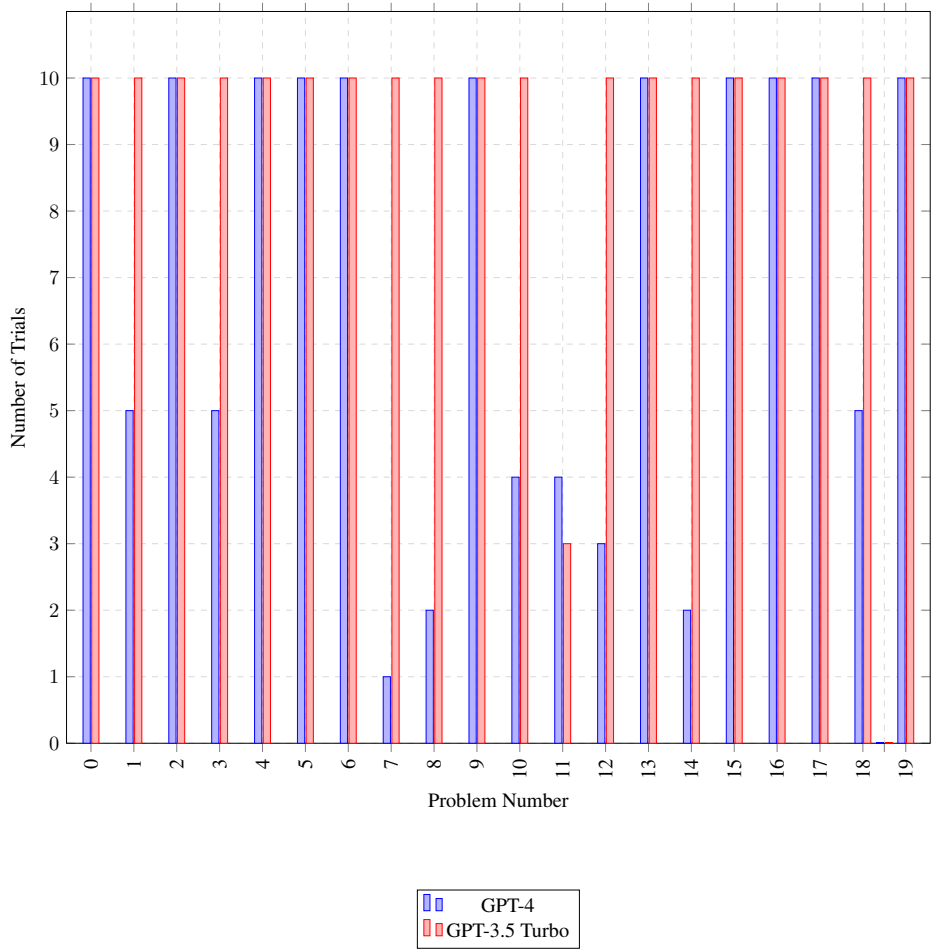



\section*{Supplementary material}

Here is the Python Z3 API code that describes the logic of the unstack operation as per the given statements:

```python
from z3 import *

def unstack(s1, s2, block1, block2):
    x, y = Ints('x y')
    blocks = [x, y]
    blocks_except = [b for b in blocks if b != block1 and b != block

    # Invariants for all blocks
    invariants = And(
        ForAll([x], Implies(x != block1, And(s1.table(x) == s2.table
        ForAll([x], Implies(x != block2, And(s1.table(x) == s2.table
        ForAll([x, y], Implies(And(x != block1, y != block2), s1.sta
    )

    # Block unstacked to the table
    unstack_to_table = And(
        s1.stacked(block1, block2),
        s1.clear(block1),
        s1.handsfree(),
        Not(s2.stacked(block1, block2)),
        s2.table(block1),
        s2.clear(block1),
        s2.clear(block2),
        s2.handsfree()
    )

    # Block unstacked to the hand
    unstack_to_hand = And(
        s1.stacked(block1, block2),
        s1.clear(block1),
        s1.handsfree(),
        Not(s2.stacked(block1, block2)),
        s2.hand(block1),
        Not(s2.clear(block1)),
        s2.clear(block2),
        Not(s2.handsfree())
    )

    return And(invariants, Or(unstack_to_table, unstack_to_hand))
```

```python
from z3 import *

def unstack(s1, s2, block1, block2):
    x, y = Ints('x y')
    blocks = [x, y]
    blocks_except = [b for b in blocks if b != block1 and b != block2]

    # Invariants for all blocks
    invariants = And(
        ForAll([x], Implies(x != block1, And(s1.table(x) == s2.table(x), s1.hand(x) == s2.hand(x),
s1.clear(x) == s2.clear(x)))),
        ForAll([x], Implies(x != block2, And(s1.table(x) == s2.table(x), s1.hand(x) == s2.hand(x),
s1.clear(x) == s2.clear(x)))),
        ForAll([x, y], Implies(And(x != block1, y != block2), s1.stacked(x, y) == s2.stacked(x, y))),
    )

    # Block unstacked to the table
    unstack_to_table = And(
        s1.stacked(block1, block2),
        s1.clear(block1),
        s1.handsfree(),
        Not(s2.stacked(block1, block2)),
        s2.table(block1),
        s2.clear(block1),
        s2.clear(block2),
        s2.handsfree()
    )

    # Block unstacked to the hand
    unstack_to_hand = And(
        s1.stacked(block1, block2),
        s1.clear(block1),
        s1.handsfree(),
        Not(s2.stacked(block1, block2)),
        s2.hand(block1),
        Not(s2.clear(block1)),
        s2.clear(block2),
        Not(s2.handsfree())
    )

    return And(invariants, Or(unstack_to_table, unstack_to_hand))
```



\end{document}